\documentclass[letterpaper, 10 pt, conference]{ieeeconf}

\IEEEoverridecommandlockouts
\usepackage{amsmath}
\usepackage{amssymb}
\usepackage{amsfonts}
\usepackage{booktabs}
\usepackage{graphicx}
\usepackage{url}
\usepackage{algorithm}
\usepackage{algorithmic}

\title{\LARGE \bf
Safe Remediation as Risk-Constrained Intervention Decision in Microservice Systems
}

\author{Chengxiao Dai$^{1}$, Zhaokun Yan$^{3}$, Chenjun Lei$^{4}$, Qiao Li$^{5}$, Luyan Zhang$^{2,*}$%
\thanks{$^{*}$Corresponding author ({\tt\small zhang.luya@northeastern.edu}).}
\thanks{$^{1}$School of Computer Science, University of Sydney, Sydney, Australia.}%
\thanks{$^{2}$Khoury College of Computer Sciences, Northeastern University, Boston, USA.}%
\thanks{$^{3}$School of Computation, Information and Technology, Technical University of Munich, Munich, Germany.}%
\thanks{$^{4}$College of Business and Economics, Australian National University, Canberra, Australia.}%
\thanks{$^{5}$School of Computer Science and Technology, Fujian Normal University, Fuzhou, China.}%
}

\begin{document}

\maketitle
\thispagestyle{empty}
\pagestyle{empty}

%%%%%%%%%%%%%%%%%%%%%%%%%%%%%%%%%%%%%%%%%%%%%%%%%%%%%%%%%%%%%%%%%%%%%%%%%%%%%%%%
\begin{abstract}

In modern IT operations (IT-Ops), the cost of an incorrect repair often exceeds the cost of no action at all. Yet existing automated remediation systems are designed to \emph{generate} actions rather than to \emph{decide} whether intervention is warranted, leaving safety as an afterthought enforced by manual approval. This paper makes three contributions to close this gap: (i) we reformulate safe remediation as a risk-constrained \textit{intervention decision} problem and cast it as a Constrained Markov Decision Process (CMDP) in which the agent maximizes repair success subject to a bounded false remediation rate (FRR); (ii) we introduce a three-dimensional risk decomposition comprising blast radius, reversibility, and epistemic uncertainty, providing operators with an interpretable, per-action safety interface; and (iii) we design a context-adaptive human-in-the-loop (HITL) gate that turns escalation from a binary failsafe into a bandwidth-aware control layer responsive to on-call load and business criticality. The full policy is learned offline from historical incident logs, enabling explicit control of the expected FRR. Experiments on the Train Ticket microservice benchmark with Chaos Mesh fault injection and an RCAEval-aligned fault taxonomy show that our framework reduces FRR by 39\% while improving repair success by 2.5 points over a strong runbook baseline, and cuts on-call escalation load by 17\% relative to a fixed-threshold variant.

\end{abstract}

%%%%%%%%%%%%%%%%%%%%%%%%%%%%%%%%%%%%%%%%%%%%%%%%%%%%%%%%%%%%%%%%%%%%%%%%%%%%%%%%
\section{INTRODUCTION}

Modern cloud-native and microservice systems generate a continuous stream of incidents. Automated remediation, executing repair actions without human intervention, promises to reduce Mean Time To Recovery (MTTR) and relieve on-call engineers \cite{c1}, and recent large language models (LLMs) and agentic systems can automatically generate plausible actions from runbooks, tickets, and telemetry \cite{c2,c5}, opening a path toward autonomous operations.

However, a fundamental gap remains between \textit{generating} and \textit{executing} actions. The cost of a wrong action is highly asymmetric: a misdirected repair can cascade into broader outages, erode trust, and trigger costly rollbacks \cite{c3} (e.g., a restart on a stateful service corrupts in-flight state; a rollback on a healthy replica cuts capacity as demand spikes). This is why most pipelines remain gated by manual approval or narrow rule-based automation. Existing automated approaches instead conflate two decisions in one scoring pipeline, which candidate to consider and whether it is safe to execute, optimizing success rate while implicitly ignoring risk asymmetry. The bottleneck is not the \textit{what} but the \textit{whether} and \textit{when}: what is missing is a decision layer that explicitly reasons about risk, uncertainty, and consequence before committing to execution.

This paper argues that safe automated remediation should be optimized for \textit{intervention decision} rather than \textit{action generation}: the deployment bottleneck is not proposing candidates but deciding whether intervention is safe enough to execute under operational risk. We make three contributions that convert ``candidate generation'' into ``deployable decision-making.''

\textit{(1) Decision-centric reformulation.} We recast remediation as a risk-constrained intervention decision and formalize it as a Constrained Markov Decision Process (CMDP) whose action space spans execution, escalation, evidence-gathering, and deliberate abstention, so the policy optimizes \textit{whether} to act, not merely \textit{which} action to propose. This exposes safety as a first-class objective rather than a downstream filter.

\textit{(2) A deployable safety interface.} We introduce a three-dimensional risk representation, blast radius, reversibility, and epistemic uncertainty, that replaces an uninterpretable scalar with per-action, dimension-specific knobs mapping directly to operational concerns: how widely an action spreads, whether it can be undone, and how confident the model is.

\textit{(3) Bandwidth-aware autonomy control.} We design a context-adaptive human-in-the-loop (HITL) gate that modulates escalation on real-time on-call load, business criticality, and diagnosis confidence, converting escalation from a binary failsafe into a bandwidth-aware control layer that respects operators' finite attention.

\begin{figure*}[t]
\centering
\includegraphics[width=6.1in]{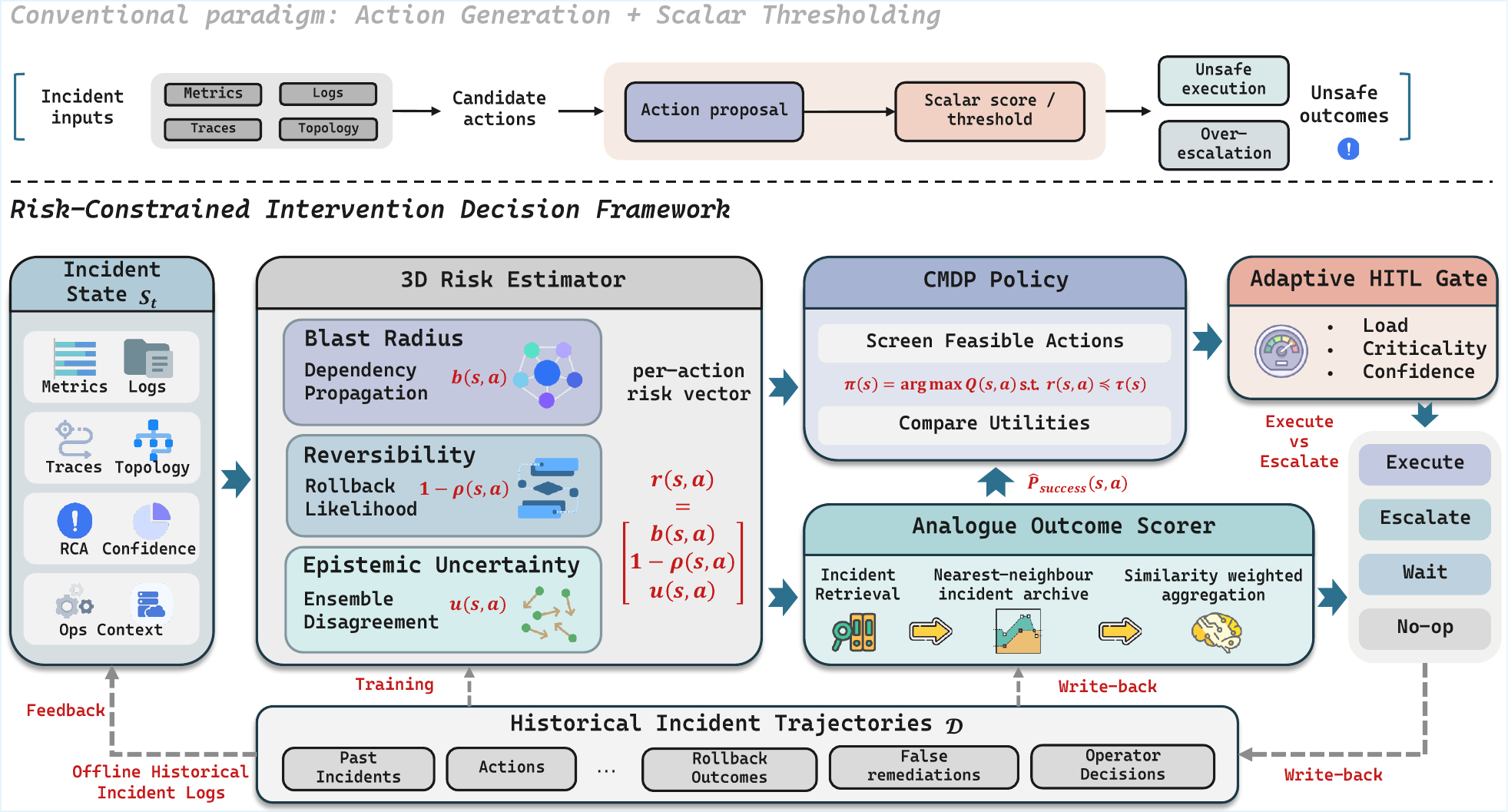}
\caption{Proposed risk-aware remediation framework.}
\label{fig:architecture}
\end{figure*}

We evaluate on the Train Ticket benchmark \cite{c17b} with Chaos Mesh \cite{c18} fault injection over an RCAEval-aligned taxonomy \cite{c19}, where our method attains the best success-safety trade-off (Sec.~V). Fig.~\ref{fig:architecture} overviews the framework.

\section{RELATED WORK}

\textit{AIOps and automated remediation.} AIOps pipelines have progressed from anomaly detection and alert correlation to root cause analysis and runbook-driven automation \cite{c1,c4}. Localization methods (MicroRCA \cite{c20}) and end-to-end frameworks (Eadro \cite{c21}) improve diagnostic accuracy on multi-source telemetry, supplying the upstream signals remediation depends on. Classical rule-engine remediation is reliable but low-coverage, while LLM generators \cite{c2,c5,c10} expand coverage; both produce \textit{candidates} and assume a downstream rule or human filters unsafe actions, the safety-aware decision layer we supply.

\textit{Safe RL and constrained MDPs.} Constrained MDPs \cite{c6} and Lagrangian methods trade reward against safety cost. CPO \cite{c11} enforces per-update constraints via trust regions but needs online interaction; offline CQL \cite{c12} penalizes out-of-distribution actions but treats safety as a soft penalty, and recent offline-RL work adds Lagrangian constraints yet models risk as one scalar budget. We decompose risk into per-dimension constraints integrated with offline Lagrangian learning for IT operations, where blast radius, reversibility, and uncertainty carry distinct semantics.

\textit{Counterfactual outcome estimation.} Causal and counterfactual policy evaluation estimate intervention outcomes in healthcare and recommendation \cite{c8,c13} but are under-explored in IT operations despite a natural fit; we use a retrieval-augmented estimator over incident analogues, capturing this spirit without a causal graph.

\textit{HITL autonomous systems.} Prior HITL frameworks use static or confidence-based escalation thresholds \cite{c9} that ignore operational context, and selective classification \cite{c14} adapts to uncertainty but not to signals such as on-call load. Our contextual bandit makes escalation a learnable function of operational state.

\textit{Risk estimation in systems.} Blast radius \cite{c15} and reversibility \cite{c16} have been studied for static dependency and configuration analysis; no prior remediation work operationalizes these dimensions within a constrained decision framework.

\section{PROBLEM FORMULATION}

\textit{System model.} We consider an IT operations environment monitored by a telemetry stack (logs, metrics, traces, alerts). The state $s_t \in \mathcal{S}$ is a feature vector aggregating: (i) the diagnosed root cause and its confidence from an upstream diagnosis module (treated as a black box \cite{c2}); (ii) affected-service summaries and topology context; (iii) telemetry anomaly features; and (iv) operational context $o_t$ (on-call load, time-of-day, business criticality).

At each decision point, the agent must select an action from the extended action set:
\begin{equation}
\mathcal{A} = \mathcal{A}_{\text{exec}} \cup \{a_{\text{escalate}}, a_{\text{wait}}, a_{\text{noop}}\},
\end{equation}
where $\mathcal{A}_{\text{exec}}$ holds candidate actions proposed upstream or retrieved from runbooks, and the remaining three denote escalating to a human, requesting more evidence, and deliberate inaction.

\textit{Constrained MDP.} We model the decision as a CMDP tuple $\langle \mathcal{S}, \mathcal{A}, P, r, c, \gamma \rangle$. The reward encodes repair success: $r = +1$ if $a$ resolves the incident, $0$ for partial resolution or escalation, and $-\kappa$ ($\kappa > 0$) for failure or worsening. The cost $c(s,a)$ indicates \textit{false remediation}: executing an action that fails or further escalates the incident. The agent seeks:
\begin{equation}
\max_{\pi} \; \mathbb{E}_\pi\!\left[\sum_t \gamma^t r(s_t, a_t) \right] \;\; \text{s.t.} \;\; \mathbb{E}_\pi\!\left[\sum_t \gamma^t c(s_t, a_t) \right] \leq \epsilon_{\text{safe}}.
\label{eq:cmdp}
\end{equation}

Unlike soft-penalty safe RL, (\ref{eq:cmdp}) is a hard operational requirement: the operator sets an acceptable FRR $\epsilon_{\text{safe}}$ that the policy must respect. This is a CMDP in the sense of \cite{c6}, solvable via Lagrangian relaxation.

\textit{Key challenges.} Three challenges distinguish this problem from classical CMDPs. (i) \textbf{Offline learning.} Online exploration is unsafe, since executing an untested action to learn its outcome may cause an outage; the policy must be learned entirely from logged decision data. (ii) \textbf{Structured risk.} Two actions with identical nominal success probability may carry very different operational risk: \texttt{restart\_db} and \texttt{clear\_cache} may both have 70\% expected success, yet the former has blast radius 0.70 / reversibility 0.25 and the latter 0.10 / 0.98. A scalar score cannot distinguish them; a structured representation can. (iii) \textbf{Heterogeneous action space.} Execution actions carry variable risk and reward, escalation is reliable but costs on-call bandwidth, and wait trades delay for evidence; a single scoring function cannot capture these asymmetries without explicit structure.

\section{METHOD}

The framework has four components, trained jointly on offline incident logs (Sec.~IV-E): a 3D risk estimator (IV-A), an analogue-based outcome scorer (IV-B), a risk-constrained policy (IV-C), and an adaptive HITL gate (IV-D).

\subsection{Three-Dimensional Risk Representation}

Given a candidate action $a \in \mathcal{A}_{\text{exec}}$ in state $s$, we compute a risk vector:
\begin{equation}
\mathbf{r}(s, a) = [\, b(s,a), \; 1-\rho(s,a), \; u(s,a) \,]^\top.
\end{equation}

\textit{Blast radius} $b(s,a)$. We propagate the influence of action $a$ over the service dependency graph $G=(V,E)$ via a learned diffusion kernel on graph powers:
\begin{equation}
b(s,a) = \frac{1}{|V|} \sum_{v \in V} \sigma_b\!\left( \sum_{k=0}^{K_{\text{diff}}} \alpha_k (A^k)_{v, \text{tgt}(a)} \right),
\end{equation}
where $A$ is the row-normalized adjacency matrix, $\text{tgt}(a)$ the target service, $\alpha_k \geq 0$ ($\sum_k \alpha_k = 1$) learnable weights, $K_{\text{diff}}$ the diffusion order, and $\sigma_b$ a learned monotone sigmoid. We fit $(\alpha, \sigma_b)$ by per-service binary cross-entropy against labels $z^{(j)}_v \in \{0,1\}$ marking whether service $v$ showed a health-metric deviation (latency, error-rate, or saturation) within 60\,s of action $a^{(j)}$, read from raw indicators rather than a separate anomaly detector.

\textit{Reversibility} $\rho(s,a)$. A model $\rho(s,a) \in [0,1]$ predicts the probability of successful rollback, trained on historical rollback logs where each action is labeled by whether rollback succeeded within an operational time budget, weighted by rollback latency. The risk contribution $1-\rho$ is high when rollback is unlikely or slow.

\textit{Epistemic uncertainty} $u(s,a)$. We maintain an ensemble of $M{=}5$ outcome predictors $\{f_m\}_{m=1}^M$, each bootstrap-resampled with independent initialization, where $f_m: \mathcal{S} \times \mathcal{A} \to \Delta^{2}$ outputs a distribution over the three outcome classes (\textit{resolved}, \textit{partial}, \textit{false remediation}). Disagreement is the average $L_2$ deviation from the ensemble mean:
\begin{equation}
u(s,a) = \frac{1}{M}\sum_{m=1}^M \left\| f_m(s,a) - \bar{f}(s,a) \right\|_2,
\end{equation}
with $\bar{f}(s,a) = \frac{1}{M}\sum_m f_m(s,a)$. High disagreement flags out-of-distribution states or actions, precisely where autonomous execution is most risky.

We do not collapse these dimensions into a scalar: the vector is checked componentwise against a state-dependent threshold $\boldsymbol{\tau}(s)$, and any single dimension exceeding its threshold rejects the action.

\subsection{Intervention Outcome Scoring via Incident Analogues}

We score the post-intervention success probability of applying $a$ in $s$:
\begin{equation}
\hat{p}_{\text{success}}(s, a) = \mathbb{P}\!\left( \text{resolved} \mid s, a \right).
\end{equation}

The estimator has three steps. (i) A learned encoder, trained jointly with $g_\psi$ to minimize outcome-prediction loss, maps each historical incident $i$ (diagnosed root cause, affected-service features, telemetry summary) to an embedding $e_i \in \mathbb{R}^d$. (ii) Given state $s$, we retrieve the top-$K$ incidents closest under cosine similarity, preferring the same fault category. (iii) We aggregate their predictions by similarity:
\begin{equation}
\hat{p}_{\text{success}}(s, a) = \sum_{i=1}^{K} w_i \cdot g_\psi(s, a, e_i),
\label{eq:scorer}
\end{equation}
where $w_i \propto \exp(\cos(e_s, e_i)/\tau_e)$ are softmax weights with temperature $\tau_e$, and $g_\psi: \mathcal{S} \times \mathcal{A} \times \mathbb{R}^d \to [0,1]$ is trained by binary cross-entropy on tuples $(s^{(i)}, a^{(i)}, y^{(i)})$ with observed outcome $y^{(i)} \in \{0,1\}$. This grounds action-value estimation in observed post-intervention outcomes, capturing the spirit of counterfactual evaluation without a fully specified causal graph.

\subsection{Risk-Constrained Policy}

The policy chooses among the extended action set according to:
\begin{equation}
\pi(s) = \arg\max_{a \in \mathcal{A}}\; Q(s,a) \;\; \text{s.t.} \;\; \mathbf{r}(s,a) \preceq \boldsymbol{\tau}(s),
\label{eq:policy}
\end{equation}
where $\boldsymbol{\tau}(s)$ is a state-dependent risk threshold and $\preceq$ is componentwise inequality. We compute the ranking value $Q$ as a convex combination of two complementary signals:
\begin{equation}
Q(s,a) = \eta \cdot \hat{p}_{\text{success}}(s,a) + (1-\eta)\cdot \tilde{Q}_\theta(s,a),
\label{eq:qcombo}
\end{equation}
where $\hat{p}_{\text{success}}$ is the analogue estimate (\ref{eq:scorer}) and $\tilde{Q}_\theta$ the CQL-regularized Q-function (Sec.~IV-E), min-max rescaled to $[0,1]$ on validation for a common scale; the two are complementary, retrieval recall vs.\ parametric generalization. The mixing weight in~(\ref{eq:qcombo}) is $\eta{=}0.6$, chosen by grid search over $\{0.3, 0.5, 0.6, 0.7\}$. When no execution action satisfies the constraint, the policy defaults to $a_{\text{escalate}}$ or $a_{\text{wait}}$ per the HITL gate (Sec.~IV-D).

\textit{Inference-time procedure.} Each decision proceeds in four steps: (1) compute the 3D risk vector for every candidate; (2) reject actions violating any dimension threshold; (3) select the highest-$Q$ survivor as the tentative execution choice; (4) execute it only if its utility exceeds the HITL-modulated escalation utility. Separating risk screening from value ranking keeps each stage independently auditable.

\textit{Two-layer safety control.} Safety is enforced along the average-vs.-tail axis: at \emph{training time} the scalar budget $\epsilon_{\text{safe}}$ governs the policy in expectation via Lagrangian relaxation, while at \emph{inference time} the 3D vector screens per action, rejecting unsafe outliers even when the aggregate budget is unexhausted. The layers are consistent, since the filter rejects only actions the trained policy already assigns low support, avoiding a train-deploy shift \cite{c7,c11}.

\textit{Lagrangian relaxation.} We convert (\ref{eq:cmdp}) into an unconstrained optimization:
\begin{equation}
\mathcal{L}(\theta, \lambda) = \mathbb{E}\!\left[ Q_\theta(s,a) \right] - \lambda \!\left( \mathbb{E}[c(s,a)] - \epsilon_{\text{safe}} \right),
\label{eq:lagrangian}
\end{equation}
updating $\theta$ and the Lagrange multiplier $\lambda \geq 0$ via alternating gradient ascent-descent until convergence.

\subsection{Context-Adaptive Escalation Gating}

Instead of a fixed escalation threshold, we treat the execution-vs.-escalation decision as a contextual bandit. Let $\phi(s)$ encode on-call load, business criticality, incident severity, diagnosis confidence, and service centrality. The gate outputs the estimated benefit of escalation:
\begin{equation}
\tau_{\text{esc}}(s) = \sigma\!\left( \mathbf{w}^\top \phi(s) + b_0 \right),
\end{equation}
with $\sigma$ logistic and $(\mathbf{w}, b_0)$ learned. High $\tau_{\text{esc}}$ (low confidence, high criticality) tilts toward escalation; low $\tau_{\text{esc}}$ (high load, clear diagnosis) favors execution. It is trained offline from observational labels, positive when the operator's execution caused a false remediation and negative on successful executions, reproducing the limited-bandwidth constraint without explicit counterfactual labels.

\subsection{Offline Safe Policy Learning}

We train offline from historical trajectories $\mathcal{D} = \{(s_t, a_t, r_t, c_t, s_{t+1})\}$. To avoid distribution shift from rarely-seen actions, we adopt a conservative Q-learning objective penalizing out-of-distribution actions \cite{c12}:
\begin{equation}
\mathcal{L}_{CQL}(\theta) = \mathcal{L}_{Q}(\theta) + \beta \!\left( \mathbb{E}_{a \sim \pi_\theta}[Q_\theta(s,a)] - \mathbb{E}_{a \sim \mathcal{D}}[Q_\theta(s,a)] \right).
\end{equation}
The overall training procedure is given in Algorithm \ref{alg:training}.

\begin{algorithm}[t]
\caption{Offline Risk-Constrained Policy Learning}
\label{alg:training}
\begin{algorithmic}[1]
\REQUIRE Offline dataset $\mathcal{D}$, budget $\epsilon_{\text{safe}}$, horizons $T_1, T_2$.
\STATE Train outcome ensemble $\{f_m\}$, rollback model $\rho$, diffusion kernel $(\alpha, \sigma_b)$ on $\mathcal{D}$.
\STATE Train outcome predictor $g_\psi$ on $(s^{(i)}, a^{(i)}, y^{(i)})$ from $\mathcal{D}$.
\STATE Initialize Q-function $Q_\theta$, Lagrange multiplier $\lambda \geq 0$.
\FOR{iter $= 1, \ldots, T_1$}
  \STATE Sample minibatch $B \sim \mathcal{D}$.
  \STATE Compute CQL loss $\mathcal{L}_{CQL}(\theta)$.
  \STATE Compute safety violation $\hat{c} = \mathbb{E}_B[c(s,a)] - \epsilon_{\text{safe}}$.
  \STATE $\theta \leftarrow \theta - \eta_\theta \nabla_\theta (\mathcal{L}_{CQL} + \lambda\hat{c})$.
  \STATE $\lambda \leftarrow \max(0, \lambda + \eta_\lambda \hat{c})$.
\ENDFOR
\FOR{iter $= 1, \ldots, T_2$}
  \STATE Train HITL gate $(\mathbf{w}, b_0)$ by contextual-bandit objective.
\ENDFOR
\end{algorithmic}
\end{algorithm}

\textit{Controlled behavior.} At convergence (\ref{eq:lagrangian}) drives $\lambda$ to enforce the FRR constraint on the offline distribution. Assuming (i) $\mathcal{D}$ covers the deployment state-action pairs and (ii) $Q_\theta$ is calibrated to within additive error $\xi$ on the support of $\mathcal{D}$, the deployment FRR is bounded by $\epsilon_{\text{safe}} + \xi + \zeta$, with $\zeta$ the concentration error of the importance-weighted off-policy estimator \cite{c7}. This is a behavioral-control argument, not a formal guarantee: (i) is unverifiable in practice, and the inference-time filter (\ref{eq:policy}) adds protection beyond the scalar Lagrangian.

\section{EXPERIMENTS}

\subsection{Datasets and Setup}

\textbf{System and telemetry.} We deploy Train Ticket \cite{c17b}, a 41-service microservice benchmark, on a 3-node Kubernetes v1.28 cluster (8\,vCPU, 32\,GB RAM per node) via the official Helm chart, giving a realistic gateway/business-logic/data-layer dependency graph. We inject faults with Chaos Mesh \cite{c18} across 11 categories aligned with RCAEval \cite{c19} (CPU, memory, disk I/O, socket, network delay, packet loss, network partition, pod kill, HTTP abort, clock skew, DNS error), each targeting a single service with known root cause, and collect telemetry from Prometheus (15\,s metrics), Fluentd (logs), and Jaeger (traces) over a 90\,s window per injection.

\textbf{State and actions.} Each state vector aggregates the diagnosed root cause and confidence (from an upstream RCA module trained on RCAEval data), anomaly scores of affected and neighboring services, and operational context (on-call load, time-of-day, business criticality). We define 12 remediation actions with intrinsic blast-radius/reversibility profiles plus three abstentions (\textit{escalate}, \textit{wait}, \textit{no-op}), 4--6 applicable per fault type. Each action runs in an isolated replica with full state reset, labeled \textit{resolved} (metrics recover within 60\,s), \textit{partial}, or \textit{false remediation} (fails or worsens downstream). The dataset has 8{,}320 decision records from 1{,}664 events, split 70/15/15.

\textbf{Historical trajectories and implementation.} The offline training set is generated by a suboptimal operator policy (softmax over action compatibilities at temperature 0.12, 25\% uniform exploration, escalation $p_{\text{esc}} \propto \text{load} \times (1 - \text{confidence})$), yielding a realistic mix of correct, suboptimal, and failed actions. All hyperparameters are tuned on the validation split and fixed at test: $M{=}5$, $K{=}8$, $\eta{=}0.6$, $\tau_e{=}0.1$, $\beta{=}1.0$, $K_{\text{diff}}{=}3$, $\epsilon_{\text{safe}}{=}0.10$, learning rates $10^{-4}/10^{-3}$ for $\theta/\lambda$, batch size 256.

\subsection{Baselines and Metrics}

\begin{figure}[htbp]
\centering
\includegraphics[width=2.5in]{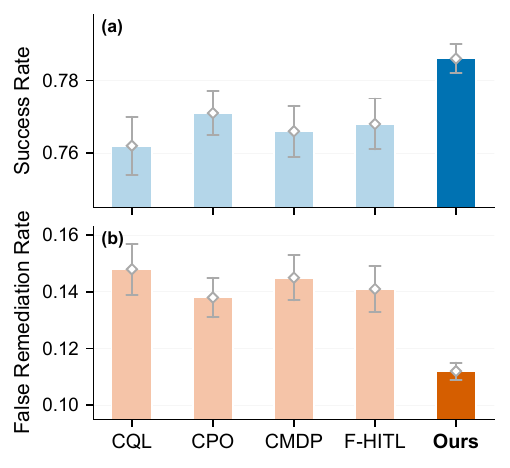}
\caption{Success (a) and FRR (b) at matched $\approx$34\% escalation. Error bars: 95\% CI, 5 seeds.}
\label{fig:matched}
\end{figure}

We compare against \textbf{Rule-Runbook} (top action from a per-fault-type runbook), \textbf{LLM-Remed} (LLM action generator without risk gating \cite{c2,c5}), \textbf{BC} (behavior cloning on operator actions), \textbf{CQL} (offline Q-learning with conservative regularization \cite{c12}, no safety constraint), \textbf{CPO (offline)} \cite{c11} with a scalar FRR constraint, and \textbf{CMDP-vanilla} (Lagrangian relaxation \cite{c6} with a single aggregate cost). Three ablations collapse risk to a scalar (\textbf{Single-Risk}), remove the risk filter (\textbf{No-Risk}), or use a fixed escalation threshold (\textbf{Fixed-HITL}). We report repair success rate, FRR, escalation rate, and time-to-recovery (TTR, including rollback cost). Success counts any resolved incident, whether autonomous or resolved by a human after escalation. We also report a matched-escalation comparison (Fig.~\ref{fig:matched}) to disentangle method quality from escalation volume.

\subsection{Main Results}

\begin{table}[h]
\caption{Main results (mean $\pm$ std, 5 seeds).}
\label{tab:main}
\begin{center}
\small
\begin{tabular}{lcccc}
\toprule
Method & Succ.\,$\uparrow$ & FRR\,$\downarrow$ & Esc.\% & TTR\,$\downarrow$ \\
\midrule
Rule-Runbook    & .761{\tiny$\pm$.005} & .183{\tiny$\pm$.007} &  0.0 & 138{\tiny$\pm$4} \\
LLM-Remed       & .724{\tiny$\pm$.008} & .214{\tiny$\pm$.009} &  0.0 & 154{\tiny$\pm$6} \\
BC              & .758{\tiny$\pm$.006} & .180{\tiny$\pm$.008} &  1.4 & 138{\tiny$\pm$5} \\
CQL             & .778{\tiny$\pm$.006} & .138{\tiny$\pm$.007} & 49.5 & 146{\tiny$\pm$5} \\
CPO (offline)   & .776{\tiny$\pm$.005} & .132{\tiny$\pm$.006} & 28.4 & 131{\tiny$\pm$4} \\
CMDP-vanilla    & .771{\tiny$\pm$.006} & .142{\tiny$\pm$.007} & 22.1 & 132{\tiny$\pm$5} \\
\midrule
Single-Risk     & .769{\tiny$\pm$.005} & .152{\tiny$\pm$.006} & 15.2 & 133{\tiny$\pm$4} \\
No-Risk         & .731{\tiny$\pm$.007} & .201{\tiny$\pm$.008} &  3.8 & 150{\tiny$\pm$6} \\
Fixed-HITL      & .779{\tiny$\pm$.005} & .124{\tiny$\pm$.006} & 43.8 & 136{\tiny$\pm$4} \\
\textbf{Ours}   & \textbf{.786}{\tiny$\pm$\textbf{.004}} & \textbf{.112}{\tiny$\pm$\textbf{.003}} & \textbf{36.5} & \textbf{126}{\tiny$\pm$\textbf{3}} \\
\bottomrule
\end{tabular}
\end{center}
\end{table}

\begin{figure}[htbp]
\centering
\includegraphics[width=2.7in]{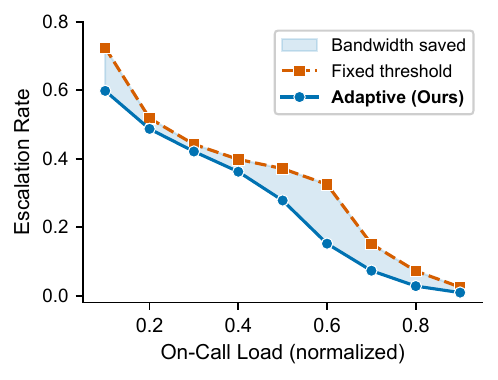}
\caption{Escalation rate vs.\ on-call load. Shaded area: bandwidth saved.}
\label{fig:escalation}
\end{figure}

At the selected operating point (Table~\ref{tab:main}), our method attains the highest success (0.786) and lowest FRR (0.112), improving success by 2.5 points over Rule-Runbook while cutting FRR by 39\%. The scalar-risk baselines CPO and CMDP-vanilla reach only FRR 0.132 and 0.142, confirming that structured decomposition gives finer safety control than a single aggregate budget, while LLM-Remed's high FRR (0.214) shows generation without a safety layer is insufficient. CQL escalates 49.5\% of decisions yet reaches FRR 0.138, and Fixed-HITL escalates 43.8\% with worse success and FRR than ours, showing that indiscriminate escalation without context-aware gating does not translate into proportional safety gains (Fig.~\ref{fig:escalation}).

\subsection{Ablation Studies}

The lower block of Table~\ref{tab:main} ablates each component. Removing the risk filter (No-Risk) raises FRR by 79\% and collapsing risk to a scalar (Single-Risk) by 36\%. Two further ablations: dropping the analogue scorer raises FRR to 0.158 ($+41\%$) and removing the CQL regularizer to 0.141 ($+26\%$); CQL is essential for down-weighting the out-of-distribution actions that behavior cloning cannot. Fixed-HITL escalates 20\% more yet scores worse on both success and FRR, confirming that adaptive gating targets escalation more effectively.

\subsection{Matched-Escalation Comparison}

To control for escalation volume, we tune each method's escalation threshold on the validation set to match $\approx$34\%, then evaluate on the test set.

\begin{table}[h]
\caption{Matched-escalation comparison ($\approx$34\%).}
\label{tab:matched}
\begin{center}
\small
\begin{tabular}{lcccc}
\toprule
Method & Esc.\% & Succ. $\uparrow$ & FRR $\downarrow$ & TTR $\downarrow$ \\
\midrule
CQL               & 34.0 & .762 & .148 & 142 \\
CPO (offline)      & 34.3 & .771 & .138 & 137 \\
CMDP-vanilla       & 33.8 & .766 & .145 & 141 \\
Fixed-HITL         & 34.6 & .768 & .141 & 139 \\
\textbf{Ours}      & \textbf{34.5} & \textbf{.784} & \textbf{.114} & \textbf{128} \\
\bottomrule
\end{tabular}
\end{center}
\end{table}

At $\sim$34\% escalation (Table~\ref{tab:matched}), our method achieves the best success (0.784) and the lowest FRR (0.114) among all competitors. Without per-dimension risk filtering, CQL and CPO reach FRR 0.148 and 0.138; Fixed-HITL reaches 0.141. This confirms that 3D risk decomposition and adaptive gating jointly yield the best use of a fixed escalation budget (Fig.~\ref{fig:matched}).

\subsection{Sensitivity and Robustness}

\textit{Escalation under load.} Fig.~\ref{fig:escalation} shows the adaptive gate cuts escalation more aggressively as on-call load grows, whereas the fixed gate barely scales down (27.8\% vs.\ 37.1\% at load 0.5; 2.8\% vs.\ 7.2\% at load 0.8, under 40\% of the fixed rate), reserving human attention for genuinely ambiguous cases under heavy load while escalating freely when bandwidth is cheap. \textit{Safety budget $\epsilon_{\text{safe}}$.} Sweeping $\epsilon_{\text{safe}} \in \{0.05, 0.10, 0.15, 0.20\}$ traces a smooth trade-off: tightening to 0.05 drives aggressive escalation (Esc.~48.2\%, Succ.~0.803, FRR~0.078), while relaxing to 0.20 executes more freely (Esc.~18.4\%, Succ.~0.768, FRR~0.162). Across the whole sweep FRR stays strictly below Rule-Runbook, so the framework tunes smoothly to operator preference; we report $\epsilon_{\text{safe}}{=}0.10$.

\textit{Retrieval size $K$.} The scorer is stable across an order of magnitude of $K \in \{2,4,8,16,32,64\}$: over $K \in [4,32]$ success and FRR vary within 0.005 and 0.008 (best at $K{=}8$: 0.786/0.112), while $K{=}2$ (0.775/0.128) over-weights outliers and $K{=}64$ (0.773/0.131) dilutes the similarity signal. We use $K{=}8$ throughout.

\textit{Upstream RCA noise.} Replacing the true root cause with a random label for a fraction $\alpha$ of test incidents (all methods re-evaluated under the same corruption), our method degrades gracefully ($\alpha{=}0.1$: 0.771/0.128; $\alpha{=}0.3$: 0.748/0.151), staying above every baseline (Rule-Runbook 0.742/0.206, CPO 0.758/0.155 at $\alpha{=}0.1$). The 3D filter acts as a second safety net: even under a wrong diagnosis, high-blast or irreversible actions are still rejected, and for novel faults with no analogues $u(s,a)$ rises sharply and triggers escalation as a zero-shot fallback.

\subsection{Case Study}

A network-partition incident on the payment service (RCA confidence 0.72, on-call load 0.45, high criticality) illustrates the full decision logic. Rule-Runbook and CQL both select \texttt{restart\_stateful}, the highest-compatibility action (0.80), which would risk corrupting in-flight transactions and cascading to six downstream services. Our policy computes its risk vector $[0.52, 0.59, 0.38]$; the irreversibility dimension (0.59) exceeds the state-dependent threshold (0.55), so the per-dimension filter rejects it despite the high score. The surviving candidate \texttt{drain\_connections} passes all dimensions ($[0.21, 0.12, 0.31]$), and its execution utility beats the HITL gate's escalation utility (0.38), so the policy executes it and the incident resolves in 52\,s. Where Fixed-HITL would escalate the entire case and consume on-call bandwidth, the three components jointly convert a risky default into a safe autonomous decision.

\section{DISCUSSION AND LIMITATIONS}

\textit{Closed action set.} The action catalog is predefined and finite. Modern remediation often requires parameterized actions such as scaling by a specific factor or rolling back to a specific commit hash; extending the framework to such spaces requires hybrid discrete-continuous policy architectures and parameter-conditioned risk estimators. \textit{Topology drift.} The blast-radius diffusion kernel is trained on a static service graph; in highly elastic environments where topologies shift on the order of minutes, the kernel may require periodic re-estimation or online adaptation.

\section{CONCLUSIONS}

This paper reframed safe automated remediation from an action-generation problem into an intervention-decision problem, exposing safety as a first-class objective and motivating three design choices: a CMDP objective with an explicit FRR budget, a three-dimensional risk decomposition that gives operators a deployable safety interface, and a context-adaptive HITL gate that turns escalation into a bandwidth-aware control layer. On a Train Ticket/Chaos Mesh benchmark with an RCAEval-aligned taxonomy, it cuts false remediation by 39\% and unnecessary escalation by 17\% over a strong fixed-threshold baseline, while remaining best across success, safety, and recovery time. The broader takeaway is that the most useful next step in autonomous remediation is not better action generation but a sharper notion of \emph{when to intervene at all}. Future work includes online safe adaptation under topology drift and parameterized action spaces.

% \addtolength{\textheight}{-12cm}

%%%%%%%%%%%%%%%%%%%%%%%%%%%%%%%%%%%%%%%%%%%%%%%%%%%%%%%%%%%%%%%%%%%%%%%%%%%%%%%%

\section*{ACKNOWLEDGMENT}

This work was supported by Chuchiang Data Co., Ltd Shanghai, China. The authors gratefully acknowledge the computational resources provided by the Chuchiang Data.

%%%%%%%%%%%%%%%%%%%%%%%%%%%%%%%%%%%%%%%%%%%%%%%%%%%%%%%%%%%%%%%%%%%%%%%%%%%%%%%%

\end{document}